# Benefits of Feature Extraction and Temporal Sequence Analysis for Video Frame Prediction: An Evaluation of Hybrid Deep Learning Models

Jose M. Sánchez Velázquez[1], Mingbo Cai[2, 3], Andrew Coney[1], Álvaro J. García-Tejedor and Alberto Nogales[1]

[1] CEIEC Research Institute, Universidad Francisco de Vitoria, Ctra. M-515 Pozuelo-Maja-dahonda km. 1.800, 28223, Pozuelo de Alarcón, Madrid, Spain
[2] Department of Psychology, University of Miami
[3] International Research Center for Neurointelligence, The University of Tokyo

alberto.nogales@ceiec.es

**Abstract.** In recent years, advances in Artificial Intelligence have significantly impacted computer science, particularly in the field of computer vision, enabling solutions to complex problems such as video frame prediction. Video frame prediction has critical applications in weather forecasting or autonomous systems and can provide technical improvements, such as video compression and streaming. Among Artificial Intelligence methods, Deep Learning has emerged as highly effective for solving vision-related tasks, although current frame prediction models still have room for enhancement. This paper evaluates several hybrid deep learning approaches that combine the feature extraction capabilities of autoencoders with temporal sequence modelling using Recurrent Neural Networks (RNNs), 3D Convolutional Neural Networks (3D CNNs), and related architectures. The proposed solutions were rigorously evaluated on three datasets that differ in terms of synthetic versus real-world scenarios and grayscale versus color imagery. Results demonstrate that the approaches perform well, with SSIM metrics increasing from 0.69 to 0.82, indicating that hybrid models utilizing 3D-CNNs and ConvLSTMs are the most effective, and greyscale videos with real data are the easiest to predict.

## 1 Introduction

Computer science has experienced significant development in recent years, largely due to the impact of new techniques in Artificial Intelligence (AI). This branch of computer science studies and interprets the mechanisms behind human intelligent behavior, aiming to replicate such behaviors in machines, though not necessarily using the same underlying processes (Hosseini et al., 2018). Although the foundations of AI were established in the 1950s, breakthroughs have only occurred in recent years, driven by the massive increase in available data and improvements in computational power.

These advancements have significantly benefited various application areas, such as signal processing, Natural Language Processing (NLP), and computer vision. The latter is defined by (Bimpas et al., 2024) as a discipline that seeks to equip computers with the ability to interpret, analyze, and understand visual information from their surroundings, including images and videos. It is within computer vision that we find one of the great milestones of modern AI: the development of AlexNet in 2012, which won the ImageNet competition with a groundbreaking performance (Krizhevsky et al., 2012). Since then, computer vision has become one of the most rapidly evolving areas of AI, with applications such as image classification, object recognition, or video frame prediction.

Video frame prediction remains a challenging task for modern AI techniques. Videos are sequences of frames over time, requiring not only image feature extraction but also sequential pattern analysis. Moreover, the task has important applications in fields like weather forecasting, video surveillance, and autonomous vehicles, as well as in technological developments such as video game design, efficient compression, and adaptive streaming. These challenges are compounded by the diversity of video content, including differences in frame nature (synthetic vs. real) and color spaces (grayscale vs. color videos). Despite the success of Deep Learning models, such as AlexNet, which is framed in the computer vision area, video frame prediction is a complex task with room for improvement. Deep Learning, as defined by (LeCun et al., 2015), involves multi-layered hierarchical artificial neural networks capable of learning data representations with progressively higher levels of abstraction from raw input.

The main motivation of this paper is to explore the effectiveness of hybrid solutions that integrate two fundamental Deep Learning models: those focused on image feature extraction and those designed for sequential pattern analysis. The performance of these hybrid approaches is evaluated in the context of video frame prediction using datasets of varying characteristics. Specifically, we test our models on three datasets: one containing grayscale synthetic data, another comprising grayscale videos of people in motion, and a third consisting of real-world color videos.

The novelty of our approach lies in predicting not the video frame itself, but its feature map, previously extracted using an Autoencoder. The same Autoencoder is then employed to reconstruct the original frame from the predicted feature map. All proposed methods follow a unified three-step structure: first, an Autoencoder extracts frame features from the input videos; second, various Deep Learning models for sequential pattern analysis, such as Recurrent Neural Networks (RNNs), are used to predict the subsequent feature map; and third, the decoder component of the Autoencoder reconstructs the video frame from the predicted features. While recent studies have utilized RNNs for direct frame prediction, to the best of our knowledge, this is the first work to evaluate the advantages of feature-level prediction in combination with Autoencoder-based reconstruction. The evaluation includes six different models tested across the three datasets, resulting in a total of 18 evaluated configurations.

This paper makes several key contributions to the field of video frame prediction. First, it introduces a novel hybrid framework that performs feature-level prediction instead of direct frame prediction, leveraging Autoencoders to both extract and reconstruct visual features, and Deep Learning–based sequential models to capture the temporal dynamics of videos. Second, the study develops six systematically designed three-stage pipelines that integrate feature extraction, sequence modelling, and frame reconstruction. Third, the proposed methods are rigorously evaluated on three diverse datasets, including synthetic grayscale, real-world grayscale, and real-world color video scenarios

The structure of this paper is as follows. Section 2 reviews prior research on video frame prediction using AI techniques. Section 3 describes the data sources and methodology employed in this study. Section 4 presents the results and their analysis, followed by a discussion of their implications in Section 5. Finally, Section 6 concludes the paper with a summary of key findings and directions for future research.

## 2 Related works

This paper aims to evaluate hybrid Deep Learning models for video frame prediction. In particular, the proposed approaches benefit from other Deep Learning models and their capabilities to extract features from images and make an analysis of sequential patterns. Related to the present work, we find different previous papers that use Artificial Intelligence techniques for video frame prediction.

Video frame prediction has been addressed for a long time, but in the case of classical Machine Learning models, this was such a complex task that it was used for intra- and inter-frame prediction. In (Kumar et al., 2013), a Machine Learning-based approach using Support Vector Machines (SVM) is developed to improve macro-block mode decision in Moving Picture Experts Group 2 (MPEG-2) video compression, specifically for inter-frame prediction. Another interesting work is (Rosa et al., 2022) that introduces GM-RF, a fast intra-frame prediction mode decision algorithm for the AO-Media Video 1 (AV1) encoder that uses Random Forest (RF) models to group and select prediction modes, significantly reducing computational complexity. (Fernandez-Escribano et al., 2008) present a low-complexity macroblock decision algorithm for transcoding MPEG-2 P-frames to H.264, focusing on inter-frame prediction using Decision Trees based on the correlation of residual data between formats. Also, (Du et al., 2015) propose a High Efficiency Video Coding (HEVC) strategy for intra-frame prediction using RF to accelerate depth decision and reduce complexity. As can be seen, classical Machine Learning models are not powerful enough to deal with frame video prediction, and all previous works are aimed at intra- or inter-frame prediction with compressed data.

With the development of Deep Learning models in recent years, the video frame prediction task has experienced great development. (Backus et al., 2022) evaluate two Deep Learning architectures, Convolutional Long Short-Term Memory (ConvLSTM) and Generative Adversarial Networks (GANs), for human motion frame prediction, training them on the UCF101 dataset. They find that the ConvLSTM model, which explicitly captures temporal and spatial patterns, outperforms GANs in predicting future frames. Another interesting work is (Desai et al., 2022), which implements a ConvLSTM model to perform next-frame prediction using realistic human activity videos from the UCF101 dataset, obtaining great results. Also, (B. Liu et al., 2021) proposes a Deep Learning framework for video compression and anomaly detection based on latent space frame prediction using ConvLSTM. Their method predicts future frames in the latent domain, achieving efficient compression and accurate anomaly detection by analyzing discrepancies between predicted and actual frame representations. In (Choi & Bajic, 2020) prediction method for video coding using a U-Net architecture is presented, performing both uni- and bi-directional inter-frame prediction without requiring motion vectors. Finally, (Straka et al., 2024) introduce PreCNet, a next-frame video prediction model implemented with hierarchical ConvLSTM modules. The model achieves state-of-the-art performance on benchmarks like KITTI and Caltech Pedestrian datasets. As we can see, Deep Learning models could achieve video frame prediction accurately, but in our case, we wonder if we could benefit from the strengths of different Deep Learning models to develop a hybrid one.

In the case of hybrid Deep Learning models for video frame prediction, we can highlight the following. (Z. Liu et al., 2021) presents a novel next-frame video prediction method that combines MobileNetV2 for frame-level feature extraction with a Transformer model for temporal sequence modelling of embeddings. Also (Mathai et al., 2024), which presents proposed HF$^2$-VAD, a hybrid video anomaly detection framework that uses Autoencoders for optical flow reconstruction and a Conditional Variational Autoencoder (CVAE) for flow-guided future frame prediction. In (Mathai et al., 2024), a hybrid video frame prediction model that combines Transformers and LSTM with 3D to efficiently model long- and short-term spatiotemporal dependencies is presented. Another interesting work is (Aravinda et al., 2025), which develops a novel hybrid video frame prediction model that fuses ConvLSTM and 3D-CNN architectures to capture temporal and spatial dynamics, respectively. As demonstrated in previous studies, many approaches combine Deep Learning models to extract spatial features from images and capture the temporal dynamics of frame sequences. In this work, we adopt a similar strategy; however, to the best of our knowledge, this is the first approach that predicts the feature map of the next frame, rather than the frame itself, which is subsequently reconstructed using an Autoencoder.

## 3    Materials and methods

In this section, we describe the datasets used in the experimentation, and then we formalize all the models that are part of the approaches.

### 3.1    Video frame datasets

As part of the paper, we are evaluating different Deep Learning model approaches. As the nature of the data could influence it, we are using three different datasets: one which contains synthetic images in black and white, another also in grayscale but with moving people, and another one which is the closest to real applications of people performing different tasks, which is colored. Following, we are providing more information about the different datasets.

The first dataset, Moving MNIST[1], consists of 10,000 sequences of 20 frames each of size 64×64 pixels, portraying two moving digits in black and white on a dark background and serving as a synthetic testbed for studying object motion and interaction in a single plane. The second dataset, ICPR'04 Recognition of Human Actions[2], comprises 599 grayscale videos with 25 subjects performing six actions: walking, jogging, running, boxing, waving, and clapping. These videos exhibit variations in scale, clothing, and setting, are recorded at 25 frames per second with a resolution of 160×120 pixels and have an average duration of four seconds. Finally, the UCF101[3] dataset contains 13,320 realistic videos sourced from YouTube, covering 101 action categories with considerable diversity in camera movement, appearance, pose, and background. Each video, at a resolution of 320×240 pixels, belongs to one of 25 groups, making the dataset a highly representative benchmark for real-world human action recognition.

## 3.2 A set of hybrid models for video frame prediction

As said before, in this work, we are evaluating different hybrid Deep Learning approaches tested in the three datasets mentioned above. The goal is to leverage the strengths of Autoencoders in extracting the essential features of images, along with the capabilities of models such as Recurrent Neural Networks (RNNs) and ConvLSTMs to analyze sequential patterns in video. Since all proposed approaches follow the same overall workflow, we describe it as follows. First, an Autoencoder is trained on all video frames to extract feature maps from its bottleneck layer, which are then used to construct a new training dataset. Second, various Deep Learning models capable of learning from time series data are trained on this feature map dataset to predict the next frame's feature representation. Finally, the decoder part of the original Autoencoder is used to reconstruct the video frame from the predicted feature map. This workflow is applied consistently across all three datasets. A schematic overview of the proposed method is presented in Fig. 1

---

[1] https://www.cs.toronto.edu/~nitish/unsupervised_video/

[2] https://www.csc.kth.se/cvap/actions/

[3] https://www.crcv.ucf.edu/data/UCF101.php

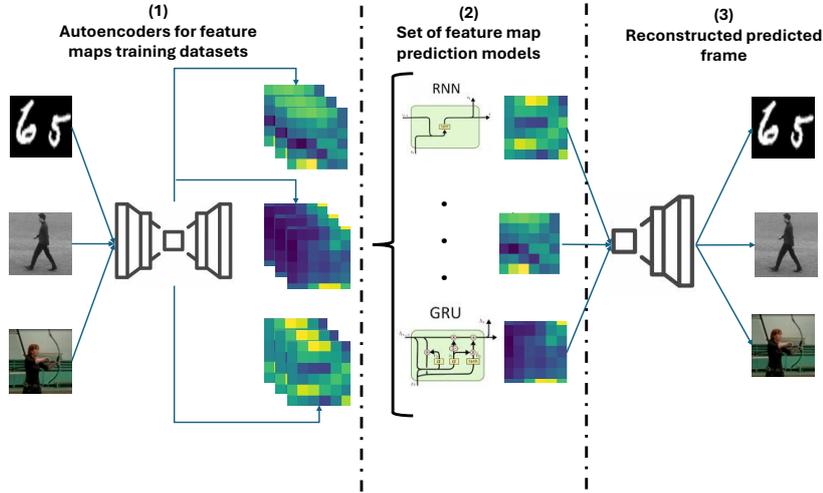

**Fig. 1.** Proposed workflow for video frame prediction based on feature maps.

As can be seen in the previous Figure, the workflow implies different Deep Learning models that are defined as follows.

**Autoencoder.** These are unsupervised learning models, training data is not labelled, originally introduced in (Ballard, 1987). A key characteristic of Autoencoders is that the input and output layers normally have the same dimensionality, and the model is composed of two main components. The first is the encoder, which compresses the input data into a lower-dimensional representation that captures its essential features. The second is the decoder, which reconstructs the original input by progressively up-sampling this compact representation back to its original size. In our case, we train an Autoencoder for each of the three datasets using pairs of the same frame. Once each Autoencoder is trained, we can obtain a dataset of feature maps to train the next models.

**Recurrent Neural Network (RNN).** First introduced by (Elman, 1990), are models designed to process input sequences of variable length by recursively applying a transition function to an internal hidden state vector, $h_t$. Due to their ability to model temporal dependencies, RNNs are particularly well-suited for analyzing time-series data, such as the present use case.

**Long Short-Term Memory (LSTM).** LSTM is a particular type of RNN introduced by (Hochreiter & Schmidhuber, 1997). The architecture features specialized components called memory blocks within the recurrent hidden layer. Each memory block includes a memory cell, which maintains the temporal state of the network through self-connections and gating mechanisms that regulate the flow of information. In the original design, each block contains an input gate and an output gate: the input gate controls how much of the incoming information is stored in the memory cell, while the output gate governs how much of the cell's activation is passed to the rest of the network.

**Gated Recurrent Unit (GRU).** This model was first described in (Cho et al., 2014) as a type of RNN designed to improve the computational efficiency of LSTM networks, especially when handling large-scale data. Regarding (Kemal Pola & Saban Öztürk, 2023), GRUs simplify the internal structure of LSTM cells by reducing the number of gates, thereby decreasing the model's time complexity. Unlike LSTMs, GRUs use only two gates: the update gate, which controls how much of the past information is carried forward, and the reset gate, which determines how much of the previous state is forgotten. This streamlined architecture enables faster training while maintaining competitive performance.

**3D Convolutional Neural Network (3D-CNN).** Introduced by (Ji et al., 2013), it is a technique used in CNNs where a 3D kernel is applied to a volume formed by stacking multiple contiguous video frames. Unlike 2D convolutions that operate only on spatial dimensions, 3D convolutions compute features across both spatial and temporal dimensions simultaneously. This allows the resulting feature maps to be influenced by information from multiple adjacent frames, enabling the network to effectively capture motion and temporal dynamics.

**Convolutional LSTM (ConvLSTM).** (Shi et al., 2015) formulate this model as an extension of the regular LSTM model. ConvLSTM is a type of RNN that incorporates convolutional structures in both the input-to-state and state-to-state transitions, allowing it to capture spatial and temporal dependencies simultaneously. When multiple ConvLSTM layers are stacked in an encoder–forecaster architecture, the model becomes well-suited for a range of spatiotemporal sequence forecasting tasks.

**Convolutional RNN (CRNN).** This model was introduced by (Ming Liang & Xiaolin Hu, 2015) to perform the task of object recognition. In (Keren & Schuller, 2017) is defined as a neural architecture that processes sequential input by feeding each frame of a sliding window into a recurrent layer. The model uses the outputs and hidden states of the recurrent units at each time step to extract temporal features across the sequence, enabling effective representation learning from sequential data

### 3.3 Performance metrics

The first metric, Mean Absolute Error (MAE), is a metric to evaluate the average magnitude of errors between the reconstructed frame and the ground truth, (Kotu & Deshpande, 2019). It is computed by taking the absolute value of the difference at each pixel to avoid cancellation of positive and negative errors and then averaging these absolute differences across all pixels. In Equation 1, the metric is formalized.

$$MSE = \frac{1}{N}\sum_{i=1}^{N} \mid y_i - \hat{y}_i \mid \qquad (1)$$

Like MAE, we also find the Mean Squared Error (MSE), which measures the Euclidean distance between the ground truth image and the image reconstructed by the Autoencoder, as described by (Sammut & Webb, 2011). A higher MSE value indicates a larger discrepancy between the reconstructed and actual image, suggesting poorer reconstructed performance. The metric is formally defined in Equation 1.

$$MSE = \frac{1}{N}\sum_{i=1}^{N}(y_i - \hat{y}_i)^2 \qquad (2)$$

For Equations 1 and 2 above, $y_i$ represents the value at a specific position in the ground truth image and $\hat{y}_i$ represents the corresponding value in the reconstructed image.

The Structural Similarity Index Measure (SSIM), introduced by (Wang et al., 2004), assesses the perceptual similarity between images, regardless of their absolute quality. It evaluates three key image components, luminance (l), contrast (c), and structure (s), each weighted by constants α, β, and γ, respectively. In Equation 2, SSIM is formalized.

$$SSIM(x,y) = [l(x,y)]^{\alpha} \cdot [c(x,y)]^{\beta} \cdot [s(x,y)]^{\gamma} \qquad (3)$$

Finally, we have the Kullback-Leibler divergence ($D_{KL}$), (Kullback & Leibler, 1951), which is defined as an asymmetric measure that quantifies how one probability distribution, $P(x)$, diverges from a baseline distribution, $Q(x)$, over the same set of values of a discrete variable $x$. We found a mathematical formalization in Equation 4.

$$D_{KL}(N(\mu_i, \sigma_i)||N(0,1)) = \frac{1}{2}\sum_i(1 + \ln(\sigma_i^2) - \mu_i^2 - \sigma_i^2) \qquad (4)$$

In the last Equation, $N(\mu,\sigma)$ is a normal distribution having as mean the mean μ and as standard deviation the standard deviation σ of the output data obtained in training, and $N(0,1)$ is the standard normal distribution.

## 4    Materials and methods

This section presents how the resulting datasets used for the evaluation have been preprocessed and how the different approaches were trained with them. It also details the metric values obtained to build the evaluated frame prediction models. Finally, a comparison between the performance of the best approaches for each of the datasets is presented.

### 4.1    Preprocessing the datasets for a fair comparison

As we have seen above, we have three datasets that differ in their frames but also in the number of videos, their length (which has consequences in the number of training frames), and the size of the frames. As we want to make a fair comparison, the following preprocessing stage has been done.

To ensure a fair comparison among the three datasets, a rigorous preprocessing stage was conducted, addressing differences in image type, number of videos, sequence lengths, and frame sizes.

First, the temporal continuity of each dataset was verified. In the case of MNIST, which was already provided as individual frames in a NumPy[4] array, geometric and centroid analyses confirmed that consecutive frames contained consistent object shapes, and that distance increased naturally with frame separation. Selected sequences were then reassembled into videos to visually confirm continuous digit motion. For the ICPR'04 and UCF101 datasets, originally provided as compressed video files, each video was decomposed into individual frames using OpenCV[5], and a sample video from each set was reconstructed to validate that the frame extraction process had preserved the correct sequence order.

Next, a common reference framework was established. All sequences were standardized to 20 frames each, using truncation for those with more frames, since MNIST originally contained 20-frame sequences. A total of 599 sequences from each dataset were then selected, matching the smallest dataset size, where MNIST was randomly reduced from its 10,000 available sequences and UCF101 was stratified by action category to ensure balanced representation. Spatial resolution was uniformly set to 64×64 pixels, which included removing black borders in certain UCF101 frames and using Lanczos interpolation for consistent resizing. MNIST frames were binarized using Otsu's method, improving contrast for white digits on a black background, while ICPR'04 frames remained grayscale and UCF101 frames remained in color. Consequently, all models trained on these datasets receive data of uniform size and length, allowing for a more direct and unbiased performance comparison.

## 4.2    Training a set of video frame prediction approaches

As described in the proposed workflow, the process can be divided into three main steps. First, three Autoencoders are individually trained to reconstruct the images from each of the three datasets. Second, various spatiotemporal models are trained on the corresponding feature maps extracted from each dataset, enabling them to predict the next feature map in the sequence of a video. Finally, the predicted feature map is passed through the corresponding Autoencoder to reconstruct the target video frame. This three-stage pipeline involves separate training processes and requires the evaluation of different metrics at each stage to assess the performance of the models. Following, we have compiled all the information related to this part of the experimentation.

**An Autoencoder for image reconstruction.** The first model that we are training for each dataset is an Autoencoder, which has two aims. First, as the present work is based on the idea of predicting feature maps, we need to construct these datasets, which are obtained from the three trained Autoencoders (one for each studied dataset). Second, this Autoencoder will be used to reconstruct the frame by using a predicted feature map obtained with the models from the second stage of our method.

---

[4] https://numpy.org/

[5] https://opencv.org/

With this in mind, we have trained three Autoencoders, one for each evaluated dataset. In all the cases, we have applied the best practices enunciated by (Nogales et al., 2024) to obtain accurate models. First, before training the models, the creation of the training, validation, and test subsets is needed. In Deep Learning, it is common practice to split the data with an 80-20 ratio between training and test sets, followed by an 80-20 split of the training set to create the validation set. This approach ensures that each subset remains representative of the overall dataset.

In our case, however, there are specific conditions that must be considered to maintain entire sequences without splitting individual videos. The 80-20 ratio was followed, yielding 480 sequences for training and 119 for testing per dataset, corresponding to 9,600 training frames and 2,380 testing frames in each. This approach prevents temporal information leaks by keeping entire sequences within one partition and ensures a balanced, reproducible data division. Apart from that, all images were normalized in ranges from 0 to 1 so they can adapt better to the used models.

Once we have our training and test subsets, we are training our Autoencoders. A grid search strategy was employed to optimize hyperparameters by exploring various combinations (Bergstra & Bengio, 2012). Table 1 summarizes the hyperparameters used across the different machine learning models during this process.

**Table 1.** Table captions should be placed above the tables.

| Hyperparameter | Values |
|---|---|
| Dimensions | [32, 64, 128], [64, 128, 256] |
| Loss function | L1, MSE, MSLE, RMSE |
| Optimizer | Adam RSMProp |
| Learning rate | 0.001, 0.0005 |

Apart from the parameters obtained through the grid search, the following hyperparameters were applied to each model. Convolutional layers were used with a kernel size of 3, a stride of 2 (halving and expanding the image dimensions), and padding set to 1 to prevent information loss at the edges. Each layer includes normalization and uses Leaky ReLU as the activation function. The weights were initialized using He Initialization, adapted for Leaky ReLU.

We have used MSE as the guidance metric to choose the model that performs the best for each dataset. This metric has been complemented with the others used in this work: MSE, MAE, SSIM, and $D_{KL}$. Table 2 compiles the MSE for training, validation, and test, for each stage.

**Table 2. MSE values for Autoencoders**

| Dataset | Training | Validation | Test |
|---|---|---|---|
| Moving MNIST | $2.78 \times 10^{-5}$ | $3.43 \times 10^{-5}$ | $4.20 \times 10^{-5}$ |
| ICPR'04 | 0.0010 | 0.0010 | 0.0015 |
| UCF101 | 0.0017 | 0.0018 | 0.0018 |

All the results presented in Table 2 indicate that the models perform well, showing no signs of overfitting or underfitting. This observation aligns with the bias-variance trade-off, as described by (Belkin et al., 2019). Regarding that the performance of the models cannot be compared with human performance, bias seems good as the metrics are very low. In terms of variance, none of the models appear to exhibit overfitting as differences between training, validation, and test are very low. We have used MSE as a guidance metric to measure the performance of the model at first, but we have complemented this with Table 3, which confirms the good performance of the three Autoencoders.

**Table 3.** Complete evaluation of Autoencoders.

| Dataset | MAE | SSIM | $D_{KL}$ |
|---------|-----|------|----------|
| Moving MNIST | 0.0014 | 0.9994 | 0.0063 |
| ICPR'04 | 0.0281 | 0.9089 | 0.0014 |
| UCF101 | 0.0313 | 0.8928 | 0.0139 |

Once we confirm that we have obtained an Autoencoder able to reconstruct the frames in each of the datasets, we compile the features of the architectures in Table 4.

**Table 4.** Architecture of Autoencoders.

| Dataset | Dimensions | Loss function | Optimizer | Learning rate |
|---------|-----------|---------------|-----------|---------------|
| Moving MNIST | [64, 128, 256] | L1 | Adam | 0.001 |
| ICPR'04 | [32, 64, 128] | RMSE | Adam | 0.005 |
| UCF101 | [64, 128, 256] | MSLE | RMSProp | 0.001 |

**A set of spatiotemporal models for predicting feature maps.** Once we have obtained the Autoencoder for each dataset, we use the feature maps of their bottlenecks to train the different spatiotemporal Deep Learning approaches. In total, we are training six models for the three obtained datasets, which makes a total of 18 solutions. In this case, the dataset was split in an 80-20 proportion into training, validation, and test subsets, and a grid search was applied for each model. In Table 5, we compile all the tuned hyperparameters and their values. We must highlight that all the models share the same hyperparameters and values, except for the number of hidden layers. In this case, 3D-CNN and RCNN do not utilize it. In this case, K-Fold validation has been applied to ensure the generalizability of the results and prevent overfitting. K-fold validation is a technique used to assess the performance of statistical models and ensure their reliability across different splits of training and test data.

**Table 5.** Grid search hyperparameters

| Hyperparameter | Values |
|---|---|
| Number of hidden layers | 1, 2, 3 |
| Size of hidden layers | 128, 256 |
| Loss function | L1, MSE, MSLE, RMSE |
| Optimizer | Adam, RMSProp |
| Learning rate | 0.01, 0.001, 0.0001 |
| Number of input frames | 3, 5, 10 |

Once we have applied the grid search for the 6 models for each dataset, the MSE has been used to choose the best approach in each of the 3 use cases. Table 6 provides the values of this metric for each dataset in training, validation, and test.

**Table 6.** MSE for the different spatiotemporal models regarding the dataset

| Dataset | Model | Training | Validation | Test |
|---|---|---|---|---|
| Moving MNIST | RNN | $0.0650 \pm 0.0003$ | $0.0651 \pm 0.0003$ | 0.0642 |
| | LSTM | $0.0510 \pm 0.0004$ | $0.0517 \pm 0.0002$ | 0.0508 |
| | GRU | $0.0562 \pm 0.0002$ | $0.0556 \pm 0.0003$ | 0.0562 |
| | **3D-CNN** | $\mathbf{0.0380 \pm 0.0013}$ | $\mathbf{0.0402 \pm 0.0004}$ | **0.0394** |
| | ConvLSTM | $0.0407 \pm 0.0003$ | $0.0408 \pm 0.0002$ | 0.0402 |
| | RCNN | $0.0454 \pm 0.0005$ | $0.0462 \pm 0.0006$ | 0.0451 |
| ICPR'04 | RNN | $0.1738 \pm 0.0086$ | $0.1739 \pm 0.0088$ | 0.1646 |
| | LSTM | $0.1255 \pm 0.0012$ | $0.1266 \pm 0.0028$ | 0.1314 |
| | GRU | $0.1547 \pm 0.0026$ | $0.1553 \pm 0.0047$ | 0.1498 |
| | 3D-CNN | $0.0541 \pm 0.0074$ | $0.0576 \pm 0.0059$ | 0.0583 |
| | **ConvLSTM** | $\mathbf{0.0471 \pm 0.0002}$ | $\mathbf{0.0479 \pm 0.0010}$ | **0.0476** |
| | RCNN | $0.0460 \pm 0.0011$ | $0.0538 \pm 0.0011$ | 0.1031 |
| UCF101 | RNN | $0.4411 \pm 0.0064$ | $0.4413 \pm 0.0091$ | 0.4356 |
| | LSTM | $0.4417 \pm 0.0003$ | $0.4419 \pm 0.0012$ | 0.4441 |
| | GRU | $0.4426 \pm 0.0003$ | $0.4427 \pm 0.0012$ | 0.4448 |
| | 3D-CNN | $0.1081 \pm 0.0061$ | $0.1171 \pm 0.0050$ | 0.1639 |
| | **ConvLSTM** | $\mathbf{0.1121 \pm 0.0005}$ | $\mathbf{0.1133 \pm 0.0023}$ | **0.1168** |
| | RCNN | $0.3303 \pm 0.0154$ | $0.3317 \pm 0.0153$ | 0.3388 |

The best models for each dataset have been bolded, so we can see that the ConvLSTM is the best for two datasets and the 3D-CNN for one. In the three use cases, the best model accomplishes the bias variance trade-off, where the MNIST and ICPR'04 have the best metrics. To complement this metric, we have compiled another metric at the test stage. In this case, regarding the previous Table, only the best model for each dataset is shown. The information can be seen in Table 6.

**Table 6.** Complete evaluation for the best spatiotemporal model

| Dataset | MAE Test |
|---------|----------|
| Moving MNIST | 0.0859 |
| ICPR'04 | 0.1255 |
| UCF101 | 0.2364 |

In the following, we describe the best models for each dataset. The best model for MNIST is the 3D-CNN with hidden layers of size 256. It uses RMSE as a loss function, Adam as an optimizer, and a learning rate of 0.01. The number of input map features is 5. In the case of ICPR'04, the best model predicting feature maps is a ConvLSTM with 2 hidden layers of size 256. It uses MSE as a loss function, Adam as an optimizer, and works with a learning rate of 0.0001. The input comprises 3 instances using a kernel of 3. In the case of UCF101, the best model is also a ConvLSTM with the same hyperparameters.

**Testing the reconstructed frames from the predicted feature maps.** At the final stage, we use the Autoencoders trained in the initial step to evaluate how accurately the images can be reconstructed from the predicted feature maps. To this end, we employed the predicted feature maps obtained from the test subsets of each dataset. These feature maps were fed into the decoder part of the corresponding Autoencoder, based on the dataset being evaluated, to reconstruct the video frames. This step allows us to assess the effectiveness of our method in predicting future frames at the feature level. Following the innovative assessment of these reduced-form representations and the selection of the optimal architecture, the reconstruction enables the calculation of the SSIM between the original frames and their reconstructed counterparts. This metric provides a quantitative evaluation of visual fidelity. In Table 7, we present the SSIM scores for each model across the three datasets, highlighting the comparative performance of the proposed approaches.

**Table 7.** Mean SSIM values of the reconstructed feature maps test set images.

| Dataset | Model | Reconstructed Feature Maps |
|---------|-------|----------------------------|
| Moving MNIST | RNN | 0.5804 |
| | LSTM | 0.6289 |
| | GRU | 0.6668 |
| | **3D-CNN** | **0.7628** |
| | ConvLSTM | 0.7518 |
| | RCNN | 0.7079 |
| ICPR'04 | RNN | 0.6133 |
| | LSTM | 0.6419 |
| | GRU | 0.6871 |
| | 3D-CNN | 0.8013 |
| | **ConvLSTM** | **0.8081** |
| | RCNN | 0.7132 |
| | RNN | 0.1618 |
| | LSTM | 0.1931 |
| | GRU | 0.1935 |
| UCF101 | 3D-CNN | 0.5483 |
| | **ConvLSTM** | **0.6981** |
| | RCNN | 0.2640 |

As we can see, once again, the best-performing model for each dataset remains the same. Notably, the ConvLSTM model achieves the highest performance on the ICPR'04 dataset, outperforming all other solutions.

But frame predictions do not always respond to the same difficulty. To study that, we have done a more in-depth analysis of how each model performs this task. First, we have divided our test set into four subsets regarding how similar the predicted frame is compared to its ground truth. As these intervals measure how good our approaches predict and reconstruct a frame, we have named them as follows. The first interval is called "excellent prediction", the second is "good prediction", the third is "fair prediction", and the last one is "poor prediction". Depending on the values of SSIM and the datasets, these intervals have different features that have been compiled in Table 8, which allows us to understand how our method performs depending on the nature of the videos.

**Table 8.** Evaluation by performance intervals for the best spatiotemporal model

| Dataset | Excellent interval | Good interval | Fair interval | Poor interval | Range width |
|---|---|---|---|---|---|
| MNIST | [0.91, 0.83]: 130 | [0.83, 0.75]: 885 | [0.75, 0.67]: 721 | [0.67, 0.59]: 49 | 0.07 |
| ICPR'04 | [0.96, 0.87]: 739 | [0.87, 0.78]: 738 | [0.78, 0.69]: 372 | [0.69, 0.60]: 174 | 0.09 |
| UCF101 | [0.84, 0.68]: 1,355 | [0.68, 0.51]: 561 | [0.51, 0.35]: 90 | [0.35, 0.19]: 17 | 0.16 |

The best model on the MNIST dataset performs very well, as the best values are around 0.90. A total of 1,015 out of 1,785 frames, which is around 50% of the tested frames, fall within excellent and good prediction intervals. Notably, only 49 frames are classified as poor. The narrow range width of 0.07 suggests high consistency and low variability in the model's predictions. This strong performance can be attributed to the relatively simple visual patterns and limited variability present in the dataset, as most of the pixels are black.

On the ICPR'04 dataset, the model also demonstrates robust performance, with 1,015 out of 2,023 frames (~50%) falling into the excellent and good categories. This is the same number of frames as in the previous dataset, but it should be highlighted that the number of frames predicted as excellent is notably higher. The remaining 546 frames are either fairly or poorly predicted. Compared to MNIST, the prediction quality is slightly more variable, as reflected by the wider range of 0.09. This indicates that while the model generalizes well to moderately complex scenarios, some frames are challenging to predict accurately.

Finally, regarding UCF101, it presents the most challenging scenario for models. Although 1,916 out of 2,023 frames (~94%) are within the top two prediction categories, which contains a much higher number of frames than the other approaches. Additionally, metrics in fair or poor intervals are very low, demonstrating that performance can be improved. The significantly broader range width of 0.16 reveals a greater degree of prediction variability, indicating that the model's performance is less stable in the presence of complex spatiotemporal patterns and background motion.

These results demonstrate that while the proposed model achieves high-quality predictions across all datasets, its performance degrades slightly as the visual and motion complexity of the input increases. The shifting distribution of prediction quality, especially the decrease in excellent predictions and increase in range width, highlights the limitations of the model when dealing with real-world, high-variability video sequences. Nonetheless, the relatively high number of good and excellent predictions, even in challenging datasets, confirms the generalizability and robustness of the model. Finally, to show real use cases of the performance of the models, we show some predictions based on the previous frames for each of the datasets. In this case, for each of the datasets, we compile 4 predictions in each Figure. These predictions correspond to the 4 upper limits of the intervals and the lower limit of the interval. For each instance, we show the previous frame, the ground truth, and the predicted one.

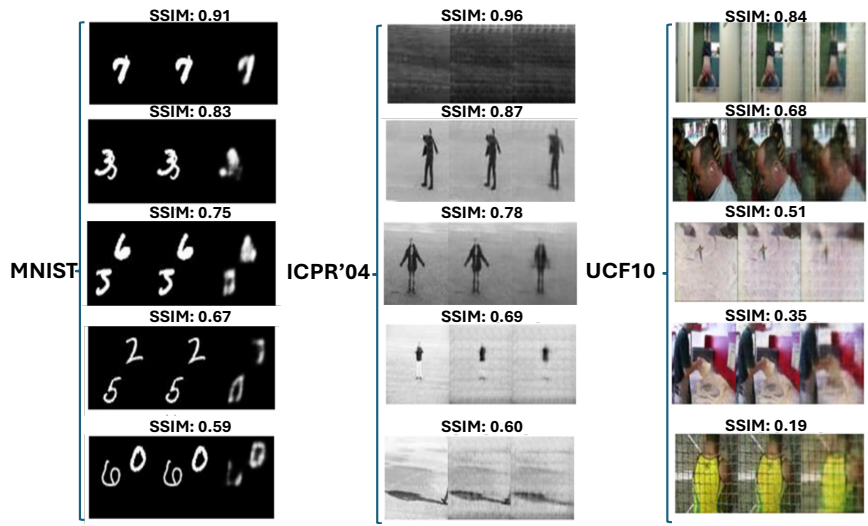

**Fig. 2.** Examples of the intervals' limits for each dataset

**Comparison against a baseline.** Regarding the previous results, our approach confirms that the feature-map-based approach not only delivers good performance in the latent space but also retains its good results once the images are restored. To assess its effectiveness in the prediction task, we compute a baseline: a spatiotemporal Deep-Learning model trained directly on the original images without leveraging feature maps.

Tables 9 (MSE) and 10 (SSIM) show the results of comparing the performance of our approaches on the three datasets against the proposed baseline.

**Table 9.** Comparison against a baseline computing MSE.

| Dataset | Our approach | Baseline |
|---|---|---|
| Moving MNIST | **0.0209** | 0.0286 |
| ICPR'04 | 0.0028 | **0.0013** |
| UCF101 | 0.0064 | **0.0028** |

**Table 10.** Comparison against a baseline computing SSIM.

| Dataset | Reconstructed Feature Maps | Baseline |
|---|---|---|
| Moving MNIST | 0.7628 | **0.8032** |
| ICPR'04 | 0.8081 | **0.9023** |
| UCF101 | 0.6981 | **0.8531** |

In Table 9 (MSE comparison), lower values indicate better performance. The proposed method outperforms the baseline on the Moving MNIST dataset (0.0209 vs. 0.0286), suggesting improved reconstruction accuracy. However, it performs worse on the ICPR'04 (0.0028 vs. 0.0013) and UCF101 (0.0064 vs. 0.0028) datasets, where the baseline achieves significantly lower error.

In Table 10 (SSIM comparison), higher values indicate better perceptual similarity. Again, the baseline achieves superior results across all datasets: Moving MNIST (0.8032 vs. 0.7628), ICPR'04 (0.9023 vs. 0.8081), and UCF101 (0.8531 vs. 0.6981). This implies that while the proposed approach may reduce pixel-wise error in some cases, it does not preserve structural and perceptual information as effectively as the baseline.

Overall, these results suggest that the proposed method has some potential, especially on the Moving MNIST dataset in terms of MSE, but it generally underperforms the baseline when it comes to perceptual fidelity and generalization to more complex datasets like UCF101 and ICPR'04.

But just having this performance comparison is not fair. As these approaches have significant practical applications, such as streaming or video compression, other aspects, such as run time or execution time. In real-time applications such as video streaming, the run time per inference iteration is critical for meeting latency constraints. Meanwhile, in offline (i.e., video compression) or batch processing scenarios, total execution time, encompassing preprocessing, model inference, and post-processing, becomes the key metric for practical deployment. In Table 11, we provide a baseline comparison in these terms.

**Table 11.** Inference analysis of the approaches.

| Dataset | Approach | Run time per iteration of predictive models | Total execution time of the approach |
|---|---|---|---|
| Moving MNIST | **3D-CNN** | **0.03 s** | **1,267 s + 1,079 s** |
| | Baseline | 0.21 s | 6,990 s |
| ICPR'04 | **ConvLSTM** | **0.06 s** | **706 s + 1,209 s** |
| | Baseline | 0.48 s | 18,892 s |
| UCF101 | **ConvLSTM** | **0.06 s** | **1,019 s + 1,038 s** |
| | Baseline | 0.32 s | 14,455 s |

Analyzing Table 11, on average, our approaches require between five and eight times less time per iteration than their baseline counterpart across all datasets. This saving is independent of sequence length, since the dimensionality reduction performed by the encoder compresses every frame into a fixed, compact representation.

Even when the one-off costs of feature extraction and final image reconstruction are added (terms shown after the plus sign), the cumulative wall-clock time decreases by factors ranging from $\approx 3 \times$ (Moving MNIST) to $\approx 10 \times$ (ICPR'04).

Another interesting analysis in DL is energy consumption due to its growing environmental impact and the escalating computational demands. As highlighted in (Aquino-Brítez et al., 2025), understanding and comparing energy efficiency across architectures is essential for promoting sustainable AI development through standardized evaluation frameworks. In Table 12, we provide such comparison of our approach against the proposed baseline.

**Table 12.** Energy-consumption analysis of the approaches.

| Dataset | Approach | Energy cost of the approach |
|---|---|---|
| Moving MNIST | **3D-CNN** | **120,160 J + 292,388 J** |
| | Baseline | 1,317,228 J |
| ICPR'04 | **ConvLSTM** | **62,628 J + 186,861 J** |
| | Baseline | 3,848,750 J |
| UCF101 | **ConvLSTM** | **94,024 J + 138,031 J** |
| | Baseline | 3,233,107 J |

Analyzing results in Table 12. Even after accounting for the encoder and decoder phases, the feature-map strategy reduces the energy budget by 70 to 94%, an advantage which increases with input complexity: ICPR'04 and UCF101, which contain richer motion and higher spatial variability, benefit the most. This significant reduction translates directly into a smaller carbon footprint and longer battery life when the models are deployed on embedded or edge devices.

Dimensionality reduction, therefore, yields substantial gains in both training time and energy consumption, which are two critical factors for continuous-inference scenarios. While the first baseline comparison sections showed that our approaches predictions incur in little worst results compared with the baseline, Tables 11 and 12 demonstrate that there is an improvement in computational sustainability which is key in the practical application of these approaches.

## 5    Conclusions and future works

This work proposes novel hybrid approaches based on Deep Learning models for video frame prediction. The core idea is to combine the feature extraction capabilities of Autoencoders with the sequential modeling power of others, such as RNNs. Based on this principle, we introduce a three-stage method: first, Autoencoders extract feature maps from the input frames; next, spatial-temporal models are trained on these feature maps to predict future frames; finally, the same Autoencoder of the first stage is used to reconstruct the future frames from the predicted feature maps.

To explore different solutions for time-series analysis, we implemented and evaluated distinct model configurations across three video datasets. Results show that the most effective models were those using ConvLSTM and 3D-CNN architectures, achieving SSIM scores of 0.72, 0.82, and 0.69 depending on the dataset. However, when compared with the baseline, our methods yielded slightly lower accuracy. In contrast, we observed that this modest trade-off in predictive performance resulted in significant gains in computational efficiency and energy consumption. This finding highlights the potential of our models for real-world applications where resource constraints are a concern.

For future work, we plan to explore the integration of advanced architectures that do not directly operate on spatiotemporal features, such as Vision Transformers and Generative Adversarial Networks (GANs). Additionally, we aim to test our models in real-world scenarios such as autonomous driving systems and first-person (Point of View) video streams to demonstrate practical applicability. Incorporating multi-modal inputs, such as optical flow, semantic segmentation maps, or audio, could further enhance the accuracy and temporal coherence of predictions, particularly in complex or dynamic scenes.

**Declarations:**

**Availability of data and materials**
Datasets are open. Code is available upon request.
**Competing interest**
The authors declare that they have no competing interests.
**Funding**
Not applicable
**Author's Contribution**
**Conceptualization** [ANM, MC] **Methodology** [ANM, MC] **Software** [AC] **Validation** [ANM, AC] **Formal Analysis** [ANM, AC, JMSV] **Investigation** [ANM, AC, JMSV] **Resources** [AGT] **Data Curation** [AC] **Writing – Original Draft** [ANM] **Writing – Review & Editing** [ANM, MC, JMSV] **Visualization** [AC] **Supervision** [ANM] **Project Administration** [AGT] **Funding Acquisition** [AGT]

**Acknowledgements**
Not applicable.

# References

Alberto Nogales, Ana M. Maitín, & Álvaro J. García-Tejedor. (2024). *Best Practices to Train Accurate Deep Learning Models: A General Methodology*.

Aquino-Brítez, S., García-Sánchez, P., Ortiz, A., & Aquino-Brítez, D. (2025). Towards an Energy Consumption Index for Deep Learning Models: A Comparative Analysis of Architectures, GPUs, and Measurement Tools. *Sensors*, *25*(3), 846. https://doi.org/10.3390/s25030846

Aravinda, C. V., Al-Shehari, T., Alsadhan, N. A., Shetty, S., Padmajadevi, G., & Reddy, K. R. U. K. (2025). A novel hybrid architecture for video frame prediction: combining convolutional LSTM and 3D CNN. *Journal of Real-Time Image Processing*, *22*(1), 50. https://doi.org/10.1007/s11554-025-01626-w

Backus, M., Jiang, Y., & Murphy, D. (2022). *Video Frame Prediction with Deep Learning*.

Ballard, D. H. (1987). Modular learning in neural networks. *AAAI*, *647*, 279–284.

Belkin, M., Hsu, D., Ma, S., & Mandal, S. (2019). Reconciling modern machine-learning practice and the classical bias--variance trade-off. *Proceedings of the National Academy of Sciences*, *116*(32), 15849–15854.

Bergstra, J., & Bengio, Y. (2012). Random search for hyper-parameter optimization. *Journal of Machine Learning Research*, *13*(2).

Bimpas, A., Violos, J., Leivadeas, A., & Varlamis, I. (2024). Leveraging pervasive computing for ambient intelligence: A survey on recent advancements, applications and open challenges. *Computer Networks*, *239*, 110156. https://doi.org/10.1016/j.comnet.2023.110156

Cho, K., Van Merriënboer, B., Gulcehre, C., Bahdanau, D., Bougares, F., Schwenk, H., & Bengio, Y. (2014). Learning phrase representations using RNN encoder-decoder for statistical machine translation. *ArXiv Preprint ArXiv:1406.1078*.

Choi, H., & Bajic, I. V. (2020). Deep Frame Prediction for Video Coding. *IEEE Transactions on Circuits and Systems for Video Technology*, 1–1. https://doi.org/10.1109/TCSVT.2019.2924657

Desai, P., Sujatha, C., Chakraborty, S., Ansuman, S., Bhandari, S., & Kardiguddi, S. (2022). Next frame prediction using ConvLSTM. *Journal of Physics: Conference Series*, *2161*(1), 012024. https://doi.org/10.1088/1742-6596/2161/1/012024

Du, B., Siu, W.-C., & Yang, X. (2015). Fast CU partition strategy for HEVC intra-frame coding using learning approach via random forests. *2015 Asia-Pacific Signal and Information Processing Association Annual Summit and Conference (APSIPA)*, 1085–1090. https://doi.org/10.1109/APSIPA.2015.7415439

Elman, J. L. (1990). Finding structure in time. *Cognitive Science*, *14*(2), 179–211.

Fernandez-Escribano, G., Kalva, H., Cuenca, P., Orozco-Barbosa, L., & Garrido, A. (2008). A Fast MB Mode Decision Algorithm for MPEG-2 to H.264 P-Frame

Transcoding. *IEEE Transactions on Circuits and Systems for Video Technology*, *18*(2), 172–185. https://doi.org/10.1109/TCSVT.2008.918115

Hochreiter, S., & Schmidhuber, J. (1997). Long short-term memory. *Neural Computation*, *9*(8), 1735–1780.

Hosseini, M.-P., Pompili, D., Elisevich, K., & Soltanian-Zadeh, H. (2018). Random ensemble learning for EEG classification. *Artificial Intelligence in Medicine*, *84*, 146–158.

Ji, S., Xu, W., Yang, M., & Yu, K. (2013). 3D Convolutional Neural Networks for Human Action Recognition. *IEEE Transactions on Pattern Analysis and Machine Intelligence*, *35*(1), 221–231. https://doi.org/10.1109/TPAMI.2012.59

Kemal Polat, & and Saban Öztürk. (2023). *Diagnostic Biomedical Signal and Image Processing Applications with Deep Learning Methods*. Elsevier. https://doi.org/10.1016/C2021-0-02190-8

Keren, G., & Schuller, B. (2017). *Convolutional RNN: an Enhanced Model for Extracting Features from Sequential Data*.

Kotu, V., & Deshpande, B. (2019). Introduction. In *Data Science* (pp. 1–18). Elsevier. https://doi.org/10.1016/B978-0-12-814761-0.00001-0

Krizhevsky, A., Sutskever, I., & Hinton, G. E. (2012). Imagenet classification with deep convolutional neural networks. *Advances in Neural Information Processing Systems*, *25*, 1097–1105.

Kullback, S., & Leibler, R. A. (1951). On Information and Sufficiency. *The Annals of Mathematical Statistics*, *22*(1), 79–86. https://doi.org/10.1214/aoms/1177729694

Kumar, V., Sharma, K. G., & Jalal, A. S. (2013). *Macro-block Mode Decision in MPEG-2 Video Compression Using Machine Learning* (pp. 149–158). https://doi.org/10.1007/978-81-322-1000-9_14

LeCun, Y., Bengio, Y., & Hinton, G. (2015). Deep learning. *Nature*, *521*(7553), 436–444.

Liu, B., Chen, Y., Liu, S., & Kim, H.-S. (2021). Deep Learning in Latent Space for Video Prediction and Compression. *2021 IEEE/CVF Conference on Computer Vision and Pattern Recognition (CVPR)*, 701–710. https://doi.org/10.1109/CVPR46437.2021.00076

Liu, Z., Nie, Y., Long, C., Zhang, Q., & Li, G. (2021). A Hybrid Video Anomaly Detection Framework via Memory-Augmented Flow Reconstruction and Flow-Guided Frame Prediction. *2021 IEEE/CVF International Conference on Computer Vision (ICCV)*, 13568–13577. https://doi.org/10.1109/ICCV48922.2021.01333

Mathai, M., Liu, Y., & Ling, N. (2024). A Hybrid Transformer-LSTM Model With 3D Separable Convolution for Video Prediction. *IEEE Access*, *12*, 39589–39602. https://doi.org/10.1109/ACCESS.2024.3375365

Ming Liang, & Xiaolin Hu. (2015). Recurrent convolutional neural network for object recognition. *2015 IEEE Conference on Computer Vision and Pattern Recognition (CVPR)*, 3367–3375. https://doi.org/10.1109/CVPR.2015.7298958

Rosa, P., Palomino, D., Porto, M., & Agostini, L. (2022). GM-RF: An AV1 Intra-Frame Fast Decision Based on Random Forest. *2022 IEEE International Conference on*

*Image Processing (ICIP)*, 3556–3560. https://doi.org/10.1109/ICIP46576.2022.9897488

Sammut, C., & Webb, G. I. (2011). *Encyclopedia of machine learning*. Springer Science \& Business Media.

Shi, X., Chen, Z., Wang, H., Yeung, D.-Y., Wong, W., & Woo, W. (2015). Convolutional LSTM Network: A Machine Learning Approach for Precipitation Nowcasting. *Proceedings of the 29th International Conference on Neural Information Processing Systems - Volume 1*, 802–810.

Straka, Z., Svoboda, T., & Hoffmann, M. (2024). PreCNet: Next-Frame Video Prediction Based on Predictive Coding. *IEEE Transactions on Neural Networks and Learning Systems*, *35*(8), 10353–10367. https://doi.org/10.1109/TNNLS.2023.3240857

Wang, Z., Bovik, A. C., Sheikh, H. R., & Simoncelli, E. P. (2004). Image quality assessment: from error visibility to structural similarity. *IEEE Transactions on Image Processing*, *13*(4), 600–612.